\pdfoutput=1

\documentclass[11pt]{article}

\usepackage[preprint]{acl}

\usepackage{times}
\usepackage{latexsym}
\usepackage{makecell}
\usepackage{graphicx}
\usepackage{hyperref}
\usepackage{multirow} 
\usepackage{float} 
\usepackage{adjustbox} 
\usepackage{verbatim} 
\usepackage{amsmath} 
\usepackage[]{xcolor}
\usepackage{bm} 

\usepackage[T1]{fontenc}

\usepackage[utf8]{inputenc}

\usepackage{microtype}

\title{AutoAugment Is What You Need: \\ Enhancing Rule-based Augmentation Methods in Low-resource Regimes}

\author{Juhwan Choi\textsuperscript{1}, Kyohoon Jin\textsuperscript{2}, Junho Lee\textsuperscript{1}, Sangmin Song\textsuperscript{1} \and Youngbin Kim\textsuperscript{1,2} \\ \\
  \textsuperscript{1}Department of Artificial Intelligence, Chung-Ang University \\
  \textsuperscript{2}Graduate School of Advanced Imaging Sciences, Multimedia and Film, Chung-Ang University \\
  \texttt{\{gold5230,fhzh123,jhjo32,s2022120859,ybkim85\}@cau.ac.kr} \\
}

\begin{document}
\maketitle
\begin{abstract}
Text data augmentation is a complex problem due to the discrete nature of sentences. Although rule-based augmentation methods are widely adopted in real-world applications because of their simplicity, they suffer from potential semantic damage. Previous researchers have suggested easy data augmentation with soft labels (softEDA), employing label smoothing to mitigate this problem. However, finding the best factor for each model and dataset is challenging; therefore, using softEDA in real-world applications is still difficult. In this paper, we propose adapting AutoAugment to solve this problem. The experimental results suggest that the proposed method can boost existing augmentation methods and that rule-based methods can enhance cutting-edge pre-trained language models.  We offer the source code.\footnote{\url{https://github.com/c-juhwan/soft-text-autoaugment}}
\end{abstract}

\section{Introduction}

Data augmentation is a regularization strategy that improves model performance expanding the data held in various ways \cite{hernandez2018data}. In the natural language processing (NLP) field, data augmentation is used in various fields to alleviate data shortages, and various augmentation methods have been proposed accordingly \cite{feng2021survey, li2022data}. For example, image data can be augmented by applying simple rules, such as flipping and rotation, to image data \cite{yang2022image}, and text data can also be augmented, by simple rules such as replacing synonyms and changing the order between words \cite{zhang2015character, wei2019eda}. In addition, a method for augmenting data by generating new text using various deep learning models has also been proposed \cite{sennrich2016improving, wu2019conditional, anaby2020not, yoo2021gpt3mix, zhou2022flipda, dai2023auggpt}. 

However, as these methods often demand training data for fine-tuning before augmentation \cite{zhang2022treemix, li2022data}, it may be challenging to apply them in a low-resource environment \cite{hu2019learning, bayer2022survey, kim2021linda}. Rule-based text data augmentation methods are less costly and easy to implement; thus, they are often used in real-world problems. Despite that, the previously proposed rule-based text data augmentation methods risk not maintaining semantic consistency with original data, which is different from image data \cite{zhao2022epida}, leading to performance degradation.
To relieve this problem, methods that perform data augmentation only through random insertion of punctuation marks have also been proposed \cite{karimi2021aeda}, but they introduce fewer variations compared to easy data augmentation (EDA).  Recently, softEDA \cite{choi2023softeda}, a method applying label smoothing \cite{szegedy2016rethinking} to the augmented data, was proposed to alleviate these drawbacks.

In softEDA, a heuristic grid search was performed for the label smoothing factor (a hyperparameter for performing label smoothing). However, the method based on a heuristic search has the following disadvantages. First, a heuristic search is expensive to execute \cite{bergstra2012random}. Second, although we found the best factor value of the grid, it may not be the global optimum. There could be a better value outside the heuristic search grid; thus, revealing the possible performance gain is difficult.

This paper proposes a method to apply AutoAugment \cite{cubuk2019autoaugment}, a technique to determine the optimal factors in the data augmentation process to alleviate the limitations of previous softEDA methods. By optimizing various arguments of softEDA, it is shown that stable and effective performance improvement is possible compared to the existing rule-based strategy with static factors. In addition, the existing softEDA experiment was conducted on an entire dataset. However, more severe overfitting occurs when the given training data are insufficient \cite{althnian2021impact}, and the scope of performance improvement is greater when additional training data are obtained from a small dataset \cite{prusa2015effect, okimura2022impact}, so data augmentation becomes increasingly crucial in this low-resource environment. Therefore, this study evaluates the proposed method under a low-resource scenario and demonstrates that the proposed method is effective even under data-scarce conditions. In addition, some existing studies have argued that simple rule-based augmentation strategies are less effective in improving the performance of pre-trained language models (PLMs) \cite{longpre2020effective, zhang2022treemix, pluvsvcec2023data}. In this study, we show that through argument optimization, it is possible to improve the performance of not only BERT \cite{devlin2019bert}, the standard PLM, but also DeBERTaV3 \cite{he2022debertav3}, a cutting-edge PLM, through rule-based data augmentation.

\section{Related Work}

Data augmentation of text is primarily performed by augmenting data according to predetermined rules \cite{zhang2015character, belinkov2018synthetic, wei2019eda, karimi2021aeda, choi2023softeda} or using various deep learning models \cite{sennrich2016improving, wu2019conditional, anaby2020not, yoo2021gpt3mix, zhou2022flipda, dai2023auggpt}.  Rule-based data augmentation methods generate new data by performing perturbation in various ways, such as replacing some of the words in a given sentence with synonyms \cite{zhang2015character} or inserting typos at the character  level \cite{belinkov2018synthetic}. The easy data augmentation (EDA) \cite{wei2019eda} technique is a representative rule-based data augmentation method consisting of synonym replacement (SR), random insertion (RI), random swap (RS), and random deletion (RD). However, because such random changes can lead to the loss of semantic consistency, the “an easier data augmentation” (AEDA) technique \cite{karimi2021aeda} consisting only of the RI of six punctuation marks has also been proposed. The softEDA \cite{choi2023softeda} method compensates for the semantic damage caused by EDA by applying label smoothing to the augmented data.

Model-based augmentation methods employ  deep learning models to generate new data. Back-translation \cite{sennrich2016improving} is one of the early model-based methods. It first translates the given data into another language and back-translates it to the original language, generating different expressions with the same concept. Methods based on PLM have also been proposed, and C-BERT \cite{wu2019conditional}, LAMBADA \cite{anaby2020not}, and FlipDA \cite{zhou2022flipda} generate new data using BERT \cite{devlin2019bert}, GPT-2 \cite{radford2019language}, and T5 \cite{raffel2020exploring}, respectively. In addition, GPT-3 \cite{brown2020language} and ChatGPT, which are larger than these PLMs, have been proposed to generate new data \cite{yoo2021gpt3mix, dai2023auggpt}. Other researchers have introduced Mixup \cite{zhang2018mixup} strategy to the NLP field to augment text data \cite{guo2020nonlinear, sun-etal-2020-mixup, yoon2021ssmix}.

Moreover, some previous approaches have tried to apply AutoAugment for NLP. Text AutoAugment \cite{ren2021text}, the work closest to the proposed method, suggested applying AutoAugment to optimize hyperparameters for data augmentation. In addition, DND \cite{kim2021makes} incorporated various data augmentation methods and suggested optimizing two reward terms regarding the difficulty and consistency with the original data. While the proposed work uses AutoAugment to optimize augmentation hyperparameters, we also focus on optimizing label smoothing values for the original and augmented data.

\section{Method}

\subsection{Preliminaries}

The EDA \cite{wei2019eda} method comprises four aforementioned suboperations: SR, RI, RS, and RD. First, SR randomly selects several words in a given sentence and changes them into their synonyms. Second, RI selects a random word in the sentence and inserts its synonym at a random position in the sentence. Third, RS operation randomly selects two words in the sentence and changes their positions. Finally, RD removes each word from the sentence with a predefined probability.

Through these four suboperations, EDA introduces noise to the original data and generates augmented data. Each suboperation has a magnitude of perturbation. For instance, in the case of SR, a higher magnitude leads to the additional replacement of the original words with their synonyms. For each observed data pair $(\bf{x}, \bf{y})$ in the original dataset $\mathcal{D}$, where $\bf{x}$ denotes an input sentence and $\bf{y}$ represents the corresponding label value, the process of EDA can be formulated as follows:

\begin{equation}
\hat{\bf{x}} = \texttt{EDA}(\textbf{x}, p_{\texttt{EDA}}) = 
\newline
\begin{cases} 
\texttt{SR}(\textbf{x}, \alpha_{\texttt{SR}}) \\
\texttt{RI}(\textbf{x}, \alpha_{\texttt{RI}}) \\
\texttt{RS}(\textbf{x}, \alpha_{\texttt{RS}}) \\
\texttt{RD}(\textbf{x}, \alpha_{\texttt{RD}}) \\
\end{cases}
\end{equation}

where $\{\alpha_{\texttt{SR}},\alpha_{\texttt{RI}},\alpha_{\texttt{RS}},\alpha_\texttt{{RD}}\}$ denotes the magnitude of each suboperation, and $p_{\texttt{EDA}} = \{p_{\texttt{SR}}, p_{\texttt{RI}}, p_{\texttt{RS}}, p_{\texttt{RD}}\}$ represents the probability distribution of each suboperation to be selected, which are equal and sum to one. As indicated, EDA only modifies $\bf{x}$, and the label of augmented data is the same as for $\bf{y}$.

The softEDA \cite{choi2023softeda} is a technique that incorporates noise into the label of augmented data through label smoothing \cite{szegedy2016rethinking}. While softEDA follows the previous EDA to augment $\hat{\textbf{x}}$, the following equation defines the process of softEDA, generating a label for augmented data $\hat{\textbf{y}}$:

\begin{equation}
\begin{aligned}
    \hat{\textbf{y}} & = (1-\epsilon_{\textit{aug}})\textbf{y} + \frac{\epsilon_{\textit{aug}}}{N_{\textit{Class}}} \\
    & = 
    \begin{cases}
        (1-\epsilon_{\textit{aug}}) + \frac{\epsilon_{\textit{aug}}}{N_{\textit{Class}}} & \text{if } y = y_i \\
        \frac{\epsilon_{\textit{aug}}}{N_{\textit{Class}}} & \text{Otherwise}
    \end{cases}
\end{aligned}
\end{equation}

where $\epsilon_{\textit{aug}}$ is a smoothing factor for label smoothing.

\subsection{Proposed Method}

Previous EDA and softEDA have numerous augmentation hyperparameters and were primarily fixed or heuristically searched. This paper proposes a method to optimize these hyperparameters by adapting AutoAugment. First, we defined an augmentation policy $\mathcal{P}$ with various factors:

\begin{equation}
\begin{aligned}
\mathcal{P} = \{
p_{\textit{aug}}, p_{\texttt{SR}}, p_{\texttt{RI}}, p_{\texttt{RS}}, p_{\texttt{RD}}, \\
\alpha_{\texttt{SR}},\alpha_{\texttt{RI}},\alpha_{\texttt{RS}},\alpha_{\texttt{RD}}, \\
N_{\textit{aug}}, \epsilon_{\textit{ori}}, \epsilon_{\textit{aug}}
\}
\end{aligned}
\end{equation}

where $p_{\textit{aug}}$ indicates the probability of augmentation, $N_{\textit{aug}}$ refers to the amount of augmented data per original data point, $\epsilon_{\textit{ori}}$ represents a label smoothing factor for the original data, different from $\epsilon_{\textit{aug}}$. Following Text AutoAugment \cite{ren2021text}, we optimized the proposed policy based on sequential model-based global optimization \cite{bergstra2011algorithms}. Finding the optimal augmentation parameter for each model and dataset through this adaptation of AutoAugment with softEDA is more beneficial than inefficient grid search.

\section{Experiment}

\begin{table*}[]
\resizebox{0.98\textwidth}{!}{ 

\begin{tabular}{l|cccccccc}

\Xhline{3\arrayrulewidth}
                  & \textbf{SST2}                                                                    & \textbf{SST5}                                                                    & \textbf{CoLA}                                                                    & \textbf{SUBJ}                                                                    & \textbf{TREC}                                                                    & \textbf{MR}                                                                      & \textbf{CR}                                                                      & \textbf{PC}                                                                      \\ \hline\hline
\textbf{BERT} w/o Aug      & \begin{tabular}[c]{@{}c@{}}$80.46_{1.84}$\\ $86.08_{1.03}$\end{tabular} & \begin{tabular}[c]{@{}c@{}}$35.13_{0.74}$\\ $43.64_{0.50}$\end{tabular} & \begin{tabular}[c]{@{}c@{}}$71.49_{1.40}$\\ $75.50_{0.58}$\end{tabular} & \begin{tabular}[c]{@{}c@{}}$92.85_{0.44}$\\ $95.07_{\bm{0.22}}$\end{tabular} & \begin{tabular}[c]{@{}c@{}}$78.42_{1.30}$\\ $93.27_{0.42}$\end{tabular} & \begin{tabular}[c]{@{}c@{}}$72.11_{1.39}$\\ $81.29_{0.52}$\end{tabular} & \begin{tabular}[c]{@{}c@{}}$79.88_{0.82}$\\ $87.53_{0.60}$\end{tabular} & \begin{tabular}[c]{@{}c@{}}$88.12_{0.58}$\\ $91.15_{0.21}$\end{tabular} \\ \hline
w/ EDA            & \begin{tabular}[c]{@{}c@{}}$80.76_{1.39}$\\ $86.71_{0.63}$\end{tabular} & \begin{tabular}[c]{@{}c@{}}$36.63_{1.33}$\\ $45.08_{1.16}$\end{tabular} & \begin{tabular}[c]{@{}c@{}}$\textcolor{gray}{70.70}_{0.98}$\\ $\textcolor{gray}{73.18}_{0.52}$\end{tabular} & \begin{tabular}[c]{@{}c@{}}$93.39_{\bm{0.25}}$\\ $\textcolor{gray}{94.69}_{0.33}$\end{tabular} & \begin{tabular}[c]{@{}c@{}}$81.56_{1.71}$\\ $93.99_{1.05}$\end{tabular} & \begin{tabular}[c]{@{}c@{}}$73.18_{1.36}$\\ $\textcolor{gray}{80.41}_{0.29}$\end{tabular} & \begin{tabular}[c]{@{}c@{}}$\textcolor{gray}{79.54}_{1.15}$\\ $87.71_{0.57}$\end{tabular} & \begin{tabular}[c]{@{}c@{}}$89.64_{0.80}$\\ $\textcolor{gray}{90.81}_{0.40}$\end{tabular} \\ \hline
w/ AEDA           & \begin{tabular}[c]{@{}c@{}}$80.96_{1.63}$\\ $86.66_{0.63}$\end{tabular} & \begin{tabular}[c]{@{}c@{}}$36.54_{0.97}$\\ $44.53_{1.02}$\end{tabular} & \begin{tabular}[c]{@{}c@{}}$72.24_{1.85}$\\ $\textcolor{gray}{74.44}_{0.41}$\end{tabular} & \begin{tabular}[c]{@{}c@{}}$93.29_{0.23}$\\ $\textcolor{gray}{94.60}_{0.48}$\end{tabular} & \begin{tabular}[c]{@{}c@{}}$81.27_{2.19}$\\ $93.87_{0.75}$\end{tabular} & \begin{tabular}[c]{@{}c@{}}$74.37_{2.84}$\\ $81.57_{\bm{0.15}}$\end{tabular} & \begin{tabular}[c]{@{}c@{}}$80.67_{1.64}$\\ $87.66_{0.55}$\end{tabular} & \begin{tabular}[c]{@{}c@{}}$88.75_{0.90}$\\ $\textcolor{gray}{91.03}_{0.31}$\end{tabular} \\ \hline
w/ softEDA        & \begin{tabular}[c]{@{}c@{}}$80.80_{3.22}$\\ $87.84_{0.65}$\end{tabular} & \begin{tabular}[c]{@{}c@{}}$37.13_{1.60}$\\ $45.04_{1.28}$\end{tabular} & \begin{tabular}[c]{@{}c@{}}$72.41_{0.95}$\\ $\textcolor{gray}{74.16}_{0.99}$\end{tabular} & \begin{tabular}[c]{@{}c@{}}$93.24_{0.40}$\\ $\textcolor{gray}{94.85}_{0.39}$\end{tabular} & \begin{tabular}[c]{@{}c@{}}$82.92_{1.70}$\\ $94.68_{0.51}$\end{tabular} & \begin{tabular}[c]{@{}c@{}}$74.40_{1.27}$\\ $\textcolor{gray}{81.16}_{0.88}$\end{tabular} & \begin{tabular}[c]{@{}c@{}}$\textcolor{gray}{78.95}_{2.65}$\\ $87.94_{0.85}$\end{tabular} & \begin{tabular}[c]{@{}c@{}}$88.82_{1.63}$\\ $\textcolor{gray}{91.12}_{0.63}$\end{tabular} \\ \hline
w/ Ours           & \begin{tabular}[c]{@{}c@{}}${\bm{85.48}}_{0.57}$\\ ${\bm{88.53}}_{\bm{0.27}}$\end{tabular} & \begin{tabular}[c]{@{}c@{}}${\bm{39.88}}_{0.41}$\\ ${\bm{46.16}}_{0.63}$\end{tabular} & \begin{tabular}[c]{@{}c@{}}${\bm{74.63}}_{\bm{0.33}}$\\ ${\bm{76.66}}_{0.81}$\end{tabular} & \begin{tabular}[c]{@{}c@{}}${\bm{94.10}}_{0.35}$\\ ${\bm{95.54}}_{0.33}$\end{tabular} & \begin{tabular}[c]{@{}c@{}}${\bm{85.88}}_{\bm{1.06}}$\\ ${\bm{95.17}}_{0.54}$\end{tabular} & \begin{tabular}[c]{@{}c@{}}${\bm{79.32}}_{\bm{0.37}}$\\ ${\bm{83.10}}_{0.34}$\end{tabular} & \begin{tabular}[c]{@{}c@{}}${\bm{86.49}}_{\bm{0.22}}$\\ ${\bm{89.98}}_{\bm{0.25}}$\end{tabular} & \begin{tabular}[c]{@{}c@{}}${\bm{91.54}}_{\bm{0.11}}$\\ ${\bm{92.16}}_{0.19}$\end{tabular} \\ \hline
w/ Ours w/o LS    & \begin{tabular}[c]{@{}c@{}}$84.71_{\bm{0.44}}$\\ $88.13_{0.48}$\end{tabular} & \begin{tabular}[c]{@{}c@{}}$39.22_{\bm{0.38}}$\\ $45.45_{\bm{0.39}}$\end{tabular} & \begin{tabular}[c]{@{}c@{}}$73.80_{0.79}$\\ $76.30_{\bm{0.34}}$\end{tabular}   & \begin{tabular}[c]{@{}c@{}}$93.71_{0.35}$\\ $95.15_{\bm{0.22}}$\end{tabular} & \begin{tabular}[c]{@{}c@{}}$84.85_{1.40}$\\ $94.70_{\bm{0.46}}$\end{tabular} & \begin{tabular}[c]{@{}c@{}}$77.86_{0.53}$\\ $82.19_{0.60}$\end{tabular} & \begin{tabular}[c]{@{}c@{}}$85.70_{0.88}$\\ $89.66_{0.35}$\end{tabular} & \begin{tabular}[c]{@{}c@{}}$91.13_{0.19}$\\ $91.98_{\bm{0.18}}$\end{tabular} \\ \hline\hline

\textbf{DeBERTaV3} w/o Aug & \begin{tabular}[c]{@{}c@{}}$88.36_{0.36}$\\ $92.59_{0.73}$\end{tabular} & \begin{tabular}[c]{@{}c@{}}$35.95_{1.69}$\\ $48.77_{1.52}$\end{tabular} & \begin{tabular}[c]{@{}c@{}}$72.62_{4.24}$\\ $82.21_{0.82}$\end{tabular} & \begin{tabular}[c]{@{}c@{}}$92.23_{0.24}$\\ $94.66_{0.22}$\end{tabular} & \begin{tabular}[c]{@{}c@{}}$80.19_{3.23}$\\ $94.06_{0.43}$\end{tabular} & \begin{tabular}[c]{@{}c@{}}$82.84_{0.39}$\\ $86.22_{0.37}$\end{tabular} & \begin{tabular}[c]{@{}c@{}}$85.61_{1.20}$\\ $91.40_{0.36}$\end{tabular} & \begin{tabular}[c]{@{}c@{}}$91.22_{0.43}$\\ $91.85_{0.26}$\end{tabular} \\ \hline

w/ EDA            & \begin{tabular}[c]{@{}c@{}}$\textcolor{gray}{86.61}_{0.70}$\\ $93.25_{0.55}$\end{tabular} & \begin{tabular}[c]{@{}c@{}}$37.64_{1.23}$\\ $49.04_{0.78}$\end{tabular} & \begin{tabular}[c]{@{}c@{}}$74.83_{1.10}$\\ $\textcolor{gray}{79.24}_{0.66}$\end{tabular} & \begin{tabular}[c]{@{}c@{}}$92.85_{0.48}$\\ $94.81_{0.53}$\end{tabular} & \begin{tabular}[c]{@{}c@{}}$83.65_{1.84}$\\ $94.33_{0.99}$\end{tabular} & \begin{tabular}[c]{@{}c@{}}$83.18_{\bm{0.32}}$\\ $86.71_{0.65}$\end{tabular} & \begin{tabular}[c]{@{}c@{}}$\textcolor{gray}{84.86}_{0.73}$\\ $\textcolor{gray}{91.24}_{0.39}$\end{tabular} & \begin{tabular}[c]{@{}c@{}}$\textcolor{gray}{90.51}_{0.47}$\\ $92.3_{0.15}$\end{tabular}  \\ \hline

w/ AEDA           & \begin{tabular}[c]{@{}c@{}}$88.44_{0.80}$\\ $\textcolor{gray}{92.54}_{0.78}$\end{tabular} & \begin{tabular}[c]{@{}c@{}}$36.87_{2.88}$\\ $49.16_{0.83}$\end{tabular} & \begin{tabular}[c]{@{}c@{}}$79.29_{0.65}$\\ $82.78_{\bm{0.40}}$\end{tabular} & \begin{tabular}[c]{@{}c@{}}$92.81_{0.47}$\\ $94.92_{0.58}$\end{tabular} & \begin{tabular}[c]{@{}c@{}}$84.17_{0.79}$\\ $94.45_{0.80}$\end{tabular} & \begin{tabular}[c]{@{}c@{}}$82.87_{0.75}$\\ $\textcolor{gray}{85.77}_{1.63}$\end{tabular} & \begin{tabular}[c]{@{}c@{}}$85.76_{1.37}$\\ $\textcolor{gray}{91.09}_{0.49}$\end{tabular} & \begin{tabular}[c]{@{}c@{}}$90.61_{0.49}$\\ $92.29_{\bm{0.11}}$\end{tabular} \\ \hline

w/ softEDA        & \begin{tabular}[c]{@{}c@{}}$88.94_{1.03}$\\ $93.12_{1.05}$\end{tabular} & \begin{tabular}[c]{@{}c@{}}$38.37_{1.65}$\\ $50.34_{1.44}$\end{tabular} & \begin{tabular}[c]{@{}c@{}}$79.40_{1.51}$\\ $\textcolor{gray}{78.97}_{1.16}$\end{tabular} & \begin{tabular}[c]{@{}c@{}}$92.90_{1.08}$\\ $94.77_{0.21}$\end{tabular} & \begin{tabular}[c]{@{}c@{}}$84.58_{1.29}$\\ $94.71_{0.69}$\end{tabular} & \begin{tabular}[c]{@{}c@{}}$83.50_{0.65}$\\ $87.02_{0.50}$\end{tabular} & \begin{tabular}[c]{@{}c@{}}$86.33_{1.65}$\\ $91.81_{0.76}$\end{tabular} & \begin{tabular}[c]{@{}c@{}}$91.28_{0.82}$\\ $92.16_{0.20}$\end{tabular} \\ \hline

w/ Ours           & \begin{tabular}[c]{@{}c@{}}${\bm{91.38}}_{0.32}$\\ ${\bm{93.94}}_{\bm{0.30}}$\end{tabular} & \begin{tabular}[c]{@{}c@{}}${\bm{42.92}}_{0.52}$\\ ${\bm{52.77}}_{\bm{0.62}}$\end{tabular} & \begin{tabular}[c]{@{}c@{}}${\bm{82.56}}_{0.51}$\\ ${\bm{84.32}}_{0.49}$\end{tabular} & \begin{tabular}[c]{@{}c@{}}${\bm{94.47}}_{0.26}$\\ ${\bm{95.29}}_{0.31}$\end{tabular} & \begin{tabular}[c]{@{}c@{}}${\bm{87.70}}_{0.90}$\\ ${\bm{94.92}}_{0.62}$\end{tabular} & \begin{tabular}[c]{@{}c@{}}${\bm{85.31}}_{0.79}$\\ ${\bm{87.96}}_{\bm{0.17}}$\end{tabular} & \begin{tabular}[c]{@{}c@{}}${\bm{89.95}}_{\bm{0.51}}$\\ ${\bm{92.46}}_{\bm{0.18}}$\end{tabular} & \begin{tabular}[c]{@{}c@{}}${\bm{92.32}}_{\bm{0.19}}$\\ ${\bm{92.72}}_{0.40}$\end{tabular} \\ \hline

w/ Ours w/o LS    & \begin{tabular}[c]{@{}c@{}}$90.47_{\bm{0.26}}$\\ $93.40_{0.58}$\end{tabular} & \begin{tabular}[c]{@{}c@{}}$42.44_{\bm{0.49}}$\\ $52.54_{0.66}$\end{tabular} & \begin{tabular}[c]{@{}c@{}}$82.10_{\bm{0.43}}$\\ $83.67_{0.86}$\end{tabular} & \begin{tabular}[c]{@{}c@{}}$94.22_{\bm{0.15}}$\\ $95.15_{\bm{0.12}}$\end{tabular} & \begin{tabular}[c]{@{}c@{}}$86.57_{\bm{0.61}}$\\ $94.92_{\bm{0.18}}$\end{tabular} & \begin{tabular}[c]{@{}c@{}}$85.07_{0.58}$\\ $87.41_{0.37}$\end{tabular} & \begin{tabular}[c]{@{}c@{}}$89.47_{0.67}$\\ $92.28_{0.27}$\end{tabular} & \begin{tabular}[c]{@{}c@{}}$92.22_{0.21}$\\ $92.49_{0.33}$\end{tabular} \\ \Xhline{3\arrayrulewidth}

\end{tabular}
}
\caption{Experimental results. Each experiment has been repeated five times and the statistics are presented in $\textit{mean}_{\textit{std}}$ format. The upper side of each column denotes the results when $N_{\textit{Train}}=100$, and the lower side shows the results when $N_{\textit{Train}}=500$. The best mean and standard deviation values for each model and dataset are boldfaced. Results that reported a lower mean value than the baseline are gray.}
\label{tab-main}
\end{table*}

\subsection{Datasets and Low-resource Setting}

Eight text classification datasets were used to evaluate the proposed method. The SST2, SST5 \cite{socher2013recursive} and MR \cite{pang-etal-2002-thumbs} sentiment classification tasks are from movie reviews. The CoLA \cite{warstadt2019neural} binary classification dataset measures the linguistic acceptability of a given sentence. The SUBJ \cite{pang2004sentimental} binary classification dataset deals with the subjectivity of a sentence. PC \cite{ganapathibhotla-liu-2008-mining}, and CR \cite{hu2004mining, qian2015automated} are datasets constructed from customer reviews. In addition, the TREC \cite{li2002learning} multiclass text classification dataset is about the question type of given text. Dataset specifications can be found in Appendix~\ref{sec:appendix-dataset}.

Data augmentation becomes more important when the given data is deficient than when sufficient data can be accessed \cite{chen2023empirical}. To simulate a more challenging scenario, we evaluated the proposed method with only 100 and 500 randomly selected original data from each dataset.

\subsection{Baselines}

To validate the claim that hyperparameter optimization for the augmentation method is effective in enhancing model performance, we compared our approach with previous rule-based data augmentation methods with fixed hyperparameters. We compared the proposed method against the previous EDA, AEDA, and softEDA methods with fixed hyperparameters.

Recent studies suggest that simple rule-based augmentation methods are insufficient to enhance PLM-based models \cite{longpre2020effective, zhang2022treemix, pluvsvcec2023data}. In addition, validating the newly proposed augmentation method using cutting-edge models, not just models like BERT, is necessary \cite{zhou2022flipda}. Therefore, we adopted BERT and DeBERTaV3 \cite{he2022debertav3}, an improvement of DeBERTa \cite{he2020deberta} as the baseline model for evaluation.

\subsection{Main Results}

Table~\ref{tab-main} reports the experimental results. Previously proposed augmentation methods have faced marginal gain, or even performance degradation. Especially, softEDA has a high standard deviation compared to other methods, indicating that softEDA has difficulty being effective within a single fixed hyperparameter and requires optimization for hyperparameters. Whereas, the proposed method exhibits a stable and remarkable performance improvement within every setting, including those where other methods had performance degradation or marginal gains. This finding suggests enhancing extensive and cutting-edge PLMs with simple augmentation methods is achievable under the carefully designed data augmentation policy and hyperparameter optimization strategy. Furthermore, it is shown that our strategy has remarkably low standard deviation values compared to other techniques, showcasing that our approach is robust against statistical differences and valuable for practical application in low-resource text classification problems.

\subsection{Ablation Study}

One may wonder whether the performance improvement reported in Table~\ref{tab-main} is solely caused by the adaptation of AutoAugment, rather than the label smoothing of softEDA. To validate the effectiveness of label smoothing, we conducted an ablation study where label smoothing is not applied (i.e., $\epsilon_{\textit{ori}} = \epsilon_{\textit{aug}} = 0$). This setting is equal to optimizing only factors of EDA. ``w/ Ours w/o LS'' row of Table~\ref{tab-main} presents the experimental results, revealing that the proposed method without label smoothing is less effective than the proposed method. This finding supports that the label smoothing optimization introduced by softEDA plays a crucial role in enhancing the model.

\section{Conclusion}

This paper proposed a method to optimize various hyperparameters of rule-based text augmentation methods. The experimental results suggest that the proposed method is effective and stable, and that rule-based augmentation methods can improve cutting-edge PLMs with proper hyperparameter optimization. Future work may extend this approach to other tasks, such as natural language inference, which is more complex than the single-sentence classification conducted in this paper.

\section*{Limitations}
This paper used AutoAugment to optimize the rule-based data augmentation method. The primary weakness of AutoAugment is the computational overhead from the searching process \cite{zhang2022good}. However, under low-resource situations, where the necessity of data augmentation is emphasized, this problem can be diminished as the time consumption of the search process decreases.

\section*{Ethics Statement}
This paper proposes an optimized rule-based augmentation method. These rule-based methods are more ethically stable than model-based approaches, as the modification is performed under predefined rules. For example, back-translation can be easily exposed to the potential bias of the translation model. Methods based on PLMs also share this concern. However, rule-based augmentation methods, including the proposed method, perform modifications within a given sentence and are less likely to be exposed to unintentional bias.

\section*{Acknowledgements}
This research was supported by Basic Science Research Program through the National Research Foundation of Korea(NRF) funded by the Ministry of Education(NRF-2022R1C1C1008534), and Institute for Information \& communications Technology Planning \& Evaluation (IITP) through the Korea government (MSIT) under Grant No. 2021-0-01341 (Artificial Intelligence Graduate School Program, Chung-Ang University).

\bibliography{custom}

\appendix

\section{Dataset Specification}
\label{sec:appendix-dataset}

\begin{table}[h!]
\centering
\resizebox{0.8\columnwidth}{!}{
\begin{tabular}{c|ccc}
\Xhline{3\arrayrulewidth}
Dataset & $N_{\textit{Class}}$ & $N_{\textit{Train}}$ & $N_{\textit{Test}}$ \\ \hline\hline
SST2    & 2      & 6.9K   & 1.8K  \\
SST5    & 5      & 8.5K   & 2.2K  \\
CoLA    & 2      & 8.5K   & 0.5K  \\
SUBJ    & 6      & 8K     & 2K    \\
TREC    & 2      & 5.5K   & 0.5K  \\
MR      & 2      & 9.5K   & 1.1K  \\
CR      & 2      & 3.0K   & 0.8K  \\
PC      & 2      & 39K    & 4.5K  \\\Xhline{2\arrayrulewidth}
\end{tabular}
}
\caption{Specification of each dataset used for the experiment.}
\label{tab-dataset}
\end{table}

\begin{table*}[ht!]
\centering
\resizebox{0.8\textwidth}{!}{
\begin{tabular}{l|lllllll}
\Xhline{3\arrayrulewidth}
                & \multicolumn{1}{c}{SST2}     & \multicolumn{1}{c}{CR}       & \multicolumn{1}{c}{MR}       & \multicolumn{1}{c}{TREC}     & \multicolumn{1}{c}{SUBJ}     & \multicolumn{1}{c}{PC}       & \multicolumn{1}{c}{CoLA}     \\ \hline\hline
BERT w/o Aug    & 89.74                        & 89.08                        & 84.28                        & 95.47                        & 96.18                        & 93.44                        & 75.38                        \\
w/ EDA          & +0.71                        & {\color[HTML]{9B9B9B} -0.41} & {\color[HTML]{9B9B9B} -0.92} & +0.51                        & {\color[HTML]{9B9B9B} -0.35} & +0.58                        & {\color[HTML]{9B9B9B} -0.45} \\
w/ AEDA         & +0.22                        & +1.84                        & \textbf{+0.19}               & {\color[HTML]{9B9B9B} -0.67} & {\color[HTML]{9B9B9B} -0.30} & {\color[HTML]{9B9B9B} -0.15} & {\color[HTML]{9B9B9B} -0.34} \\
w/ softEDA 0.1  & {\color[HTML]{9B9B9B} -0.11} & +0.29                        & {\color[HTML]{9B9B9B} -1.10} & {\color[HTML]{9B9B9B} -1.45} & \textbf{+0.15}               & +0.43                        & +1.34                        \\
w/ softEDA 0.15 & {\color[HTML]{9B9B9B} -0.22} & +0.66                        & {\color[HTML]{9B9B9B} -0.46} & {\color[HTML]{9B9B9B} -0.47} & {\color[HTML]{9B9B9B} -0.50} & {\color[HTML]{9B9B9B} -0.01} & +0.02                        \\
w/ softEDA 0.2  & {\color[HTML]{9B9B9B} -0.12} & \textbf{+2.10}               & \textbf{+0.19}               & {\color[HTML]{9B9B9B} -0.27} & +0.05                        & +0.43                        & +0.81                        \\
w/ softEDA 0.25 & {\color[HTML]{9B9B9B} -0.23} & \textbf{+2.10}               & {\color[HTML]{9B9B9B} -0.92} & \textbf{+1.17}               & {\color[HTML]{9B9B9B} -0.10} & \textbf{+0.67}               & \textbf{+1.50}               \\
w/ softEDA 0.3  & \textbf{+0.83}               & {\color[HTML]{9B9B9B} -0.90} & {\color[HTML]{9B9B9B} -1.80} & {\color[HTML]{9B9B9B} -0.78} & +0.00                        & \textbf{+0.67}               & +0.23                        \\ \Xhline{3\arrayrulewidth}
\end{tabular}}
\caption{Results of softEDA for the BERT model reported in the softEDA paper. The best scores for each dataset are boldfaced. Scores lower than the baseline are gray.}
\label{tab-softeda}
\end{table*}

\section{Implementation Details}
\label{sec:appendix-detail}

We used PyTorch \cite{paszke2019pytorch} and Huggingface Transformers \cite{wolf2020transformers} to implement the model and evaluation process. We used \texttt{bert-base-cased} and \texttt{microsoft/deberta-v3-base} for the BERT and DeBERTaV3 models. Every model was trained using the Adam optimizer with a batch size of 32 and a learning rate of 5e-5 for ten epochs, with early stopping with a patience value of 5, conditioned on best validation accuracy. The training procedure was performed on a single Nvidia RTX 3090 GPU.

For the baseline method implementation, we used TextAugment library \cite{marivate2020improving} for EDA, and softEDA was built on it. The library did not have an implementation for AEDA; thus, we implemented it separately. We used ray tune \cite{liaw2018tune} to implement the proposed method. Please refer to the attached code for more information.

\section{Analysis of softEDA}
\label{sec:appendix-analysis}

We investigated the experimental results of the softEDA paper. Table~\ref{tab-softeda} presents the experimental results reported in the appendix of the softEDA paper. The results suggest that, although softEDA can potentially enhance model performance, it is problematic to determine the optimal label smoothing factor for each model and dataset. Performance degradation compared to the baseline was also observed where the factor is improper for each setup. This finding motivated us to determine a better solution for finding optimal factors than a heuristic search. Furthermore, the authors performed the experiment on the full dataset. In contrast, we conducted the experiment through low-resource scenarios, which is more challenging for model.

\end{document}